\def\paperlanguage{}
\pdfoutput=1 

\documentclass[letterpaper, 10 pt, conference]{ieeeconf}  

\usepackage{bm}
\usepackage{cite}
\usepackage{multirow}
\include{preamble}

\newcommand{\ctext}[1]{\raise0.2ex\hbox{\textcircled{\scriptsize{#1}}}}

\IEEEoverridecommandlockouts                              

\title{\LARGE \textbf
  {
    \switchlanguage%
    {%
      MEVION: Low-Cost Open-Source Data Collection System for Powerful and High-Speed Dual-Arm Manipulation
    }%
    {%
      MEVION: 力強く高速な双腕操作のための\\低コストオープンソースデータ収集システム
    }%
  }
}

\author{Kento Kawaharazuka$^{1}$, Yoshiki Obinata$^{1}$, Hirokazu Ishida$^{1}$, Jihoon Oh$^{1}$, Temma Suzuki$^{1}$, Shintaro Inoue$^{1}$,\\Keita Yoneda$^{1}$, Ayumu Iwata$^{1}$, and Kei Okada$^{1}$
  \thanks{$^{1}$ The authors are with the Department of Mechano-Informatics, Graduate School of Information Science and Technology, The University of Tokyo, 7-3-1 Hongo, Bunkyo-ku, Tokyo, 113-8656, Japan.
    {\texttt\small [kawaharazuka, obinata, h-ishida, oh, t-suzuki, s-inoue, yoneda, a-iwata, k-okada]@jsk.imi.i.u-tokyo.ac.jp}
  }%
}

\begin{document}

\maketitle
\thispagestyle{empty}
\pagestyle{empty}

\begin{abstract}
  \switchlanguage%
  {%
    The global competition for developing robotic foundation models is intensifying.
    Among the data collection systems used for dual-arm robots, ALOHA is representative of being low-cost and open-source, and is widely adopted by researchers as a de facto standard.
    However, due to its limited ability to generate high forces and speeds, it is difficult to handle heavy objects or perform fast manipulations.
    To address this, we developed MEVION, a low-cost and open-source dual-arm robot data collection system capable of generating greater force and speed.
    All parts of this robot can be sourced through e-commerce, and by extensively utilizing sheet metal welding, its large body structure is constructed with a small number of components at low cost, while also simplifying assembly.
    MEVION is equipped with four 6-DoF arms with parallel grippers.
    Each arm weighs 7.0 kg and has a maximum torque of 60 Nm, and the entire system can be constructed for about USD 14,000.
    The elbow joint adopts a closed-link mechanism similar to those used in quadruped robots, which reduces the distal mass and enables higher force and speed output at the end-effector.
    We demonstrate that MEVION enables data collection for object manipulation tasks not previously possible and supports imitation learning-based motion generation.
    All hardware and software of this work are included in the Supplementary Materials or \href{https://github.com/haraduka/mevion}{\textcolor{magenta}{github.com/haraduka/mevion}}.
  }%
  {%
    現在世界では, ロボット基盤モデルの開発競争が激化している.
    ここで用いられる双腕ロボットのデータ収集システムはALOHAが代表的であり, 低コストかつオープンソースで公開されていることから, 多くの研究者がデファクトスタンダードとして活用している.
    一方で, 大きな力と速度を出すことができないため, 重量物を扱ったり, 高速に物体を操作することは難しい.
    そこで本研究では, より大きな力と速度を出すことができる低コストかつオープンソースな双腕ロボットのデータ収集システムとして, MEVIONを開発した.
    本ロボットは全ての部品をE-Commerceにより調達可能であり, 板金溶接を駆使して大きな身体構造を少ない部品数で安価に構築, アセンブリを容易にしている.
    MEVIONは6自由度+平行グリッパのアームを4つ備え, 各アームの重量は7.0 kg, 最大トルクは60 Nmであり, 全体で約14Kドルで構築することができる.
    肘関節は四脚ロボットに見る閉リンク構造を採用しており, これにより手先の重量を軽減し, より大きな速度を出すことができる.
    MEVIONがこれまでにない多様な物体操作のデータ収集に活用可能であること, 一連の摸倣学習動作生成が可能であることを示した.
    本研究の全てのハードウェア・ソフトウェアはSupplementary Materialsに含まれている (採択後, GitHubにも公開する予定である).
  }%
\end{abstract}

\section{INTRODUCTION}\label{sec:introduction}
\switchlanguage%
{%
  Competition in the development of robotic foundation models is intensifying, with research institutes and companies worldwide vying for progress.
  In particular, the advancement of imitation learning and its extension to Vision-Language-Action (VLA) models \cite{ma2024survey, zhong2025survey, kawaharazuka2025vlasurvey} has spurred active research on manipulation using dual-arm robots.
  Various types of dual-arm robots and corresponding data collection systems have been proposed for this purpose.

  Broadly speaking, two main teaching methods exist.
  The first involves operating robots via VR devices or motion capture systems \cite{qin2023anyteleop, cheng2024open, ding2024bunny}, while the second adopts a leader-follower scheme in which humans directly move the robot's arms \cite{zhao2023aloha, wu2024gello, aldaco2024aloha2, fu2024mobile, kobayashi2025alpha, huggingface2025so101, enactic2025openarm}.
  Both approaches have their advantages and drawbacks, but the latter is generally considered more intuitive and has therefore been widely adopted by researchers.
  In terms of robot morphology, three categories can be identified: tabletop-type \cite{zhao2023aloha, wu2024gello, aldaco2024aloha2, kobayashi2025alpha, huggingface2025so101}, mobile-type \cite{fu2024mobile, cui2025aharobot}, and standing-type \cite{cheng2024open, ding2024bunny, enactic2025openarm}.
  The tabletop-types are placed on a table; mobile-types have a wheeled base; and standing-types resemble a humanoid upper body. 
  While tabletop-types and mobile-types typically have arms extending upward from below, standing-types have arms extending downward from the shoulders.
  Another important distinction is whether a system is open-source or closed-source.
  Commercially released systems are generally closed-source, making it difficult to modify link lengths or change the setup.
  In contrast, open-source systems offer the advantage of customizability.
  Considering the increasing complexity of future manipulation tasks, the ability to freely reconfigure systems through open-source design is a crucial factor.

  Within this landscape, this study focuses on tabletop-type dual-arm robot data collection systems employing the leader-follower method.
  Among these, ALOHA \cite{zhao2023aloha} is representative of being both low-cost and open-source, and has become a de facto standard widely used by researchers.
  However, since it relies on small servo motors, it cannot generate high forces or speeds, making it difficult to handle heavy objects or perform rapid manipulations.
  As a result, studies using ALOHA and similar systems have primarily focused on tasks involving slow manipulation of lightweight objects, posing challenges for collecting more diverse manipulation data.
}%
{%
  現在, ロボット基盤モデルの開発競争が激化しており, 各国の研究機関や企業がしのぎを削っている.
  特に, 摸倣学習やそれを発展させたVision-Language-Action (VLA) models \cite{ma2024survey, zhong2025survey, kawaharazuka2025vlasurvey}の発展により, 双腕ロボットを用いたマニピュレーションの研究が盛んに行われている.
  ここで用いられる双腕ロボットとそのデータ収集システムには, 様々な種類が提案されている.

  まず教示方法には, 大きく分けて, VRデバイスやモーションキャプチャを用いてロボットを操作する方法\cite{qin2023anyteleop, cheng2024open, ding2024bunny}と, leader-follower方式で人間がロボットのアームを直接動かす方法\cite{zhao2023aloha, wu2024gello, aldaco2024aloha2, fu2024mobile, kobayashi2025alpha, huggingface2025so101, enactic2025openarm}がある.
  どちらにもメリット・デメリットはあるが, 後者はより直感的な操作が可能であり, 多くの研究者が採用している.
  次にロボット形態としては, 大きく分けて, テーブルトップ型\cite{zhao2023aloha, wu2024gello, aldaco2024aloha2, kobayashi2025alpha, huggingface2025so101}, 台車型\cite{fu2024mobile, cui2025aharobot}, 立位型\cite{cheng2024open, ding2024bunny, enactic2025openarm}の3種類がある.
  テーブルトップ型はテーブル上にロボットが配置されており, 移動型はそれに台車がついており, 立位型はヒューマノイドの上半身のような形態をしている.
  テーブルトップ型と台車型では大抵腕が下から上に伸びているが, 立位型は腕が肩から下に伸びている.
  最後に, オープンソースかクローズドソースかの観点がある.
  企業から発売されているシステムはクローズドソースが一般的であるが, 一部のリンク長さを変えたり, セットアップを変えることは基本的に難しい.
  それに対して, オープンソースの場合は, 自分で自由にカスタマイズできるという利点がある.
  今後よりマニピュレーションが高度化していくことを考えると, オープンソースで自由に組み換えできるという性質は重要な要素である.

  その中でも本研究では, leader-follower方式のテーブルトップ型双腕ロボットデータ収集システムに着目した.
  この領域では, ALOHA \cite{zhao2023aloha}が代表的である.
  低コストかつオープンソースで公開されていることから, 多くの研究者のデファクトスタンダードとして活用されている.
  一方で, 小型のサーボモータを使用しているため, 大きな力と速度を出すことができず, 重量物を扱ったり, 高速に物体を操作することは難しい.
  これによって, ALOHAやそれに類似したいシステムを用いた研究では, 軽い物体をゆっくりと操作するタスクが中心となっており, より多様な物体操作のデータ収集が難しいという課題がある.
}%

\begin{figure}[t]
  \centering
  \includegraphics[width=0.9\columnwidth]{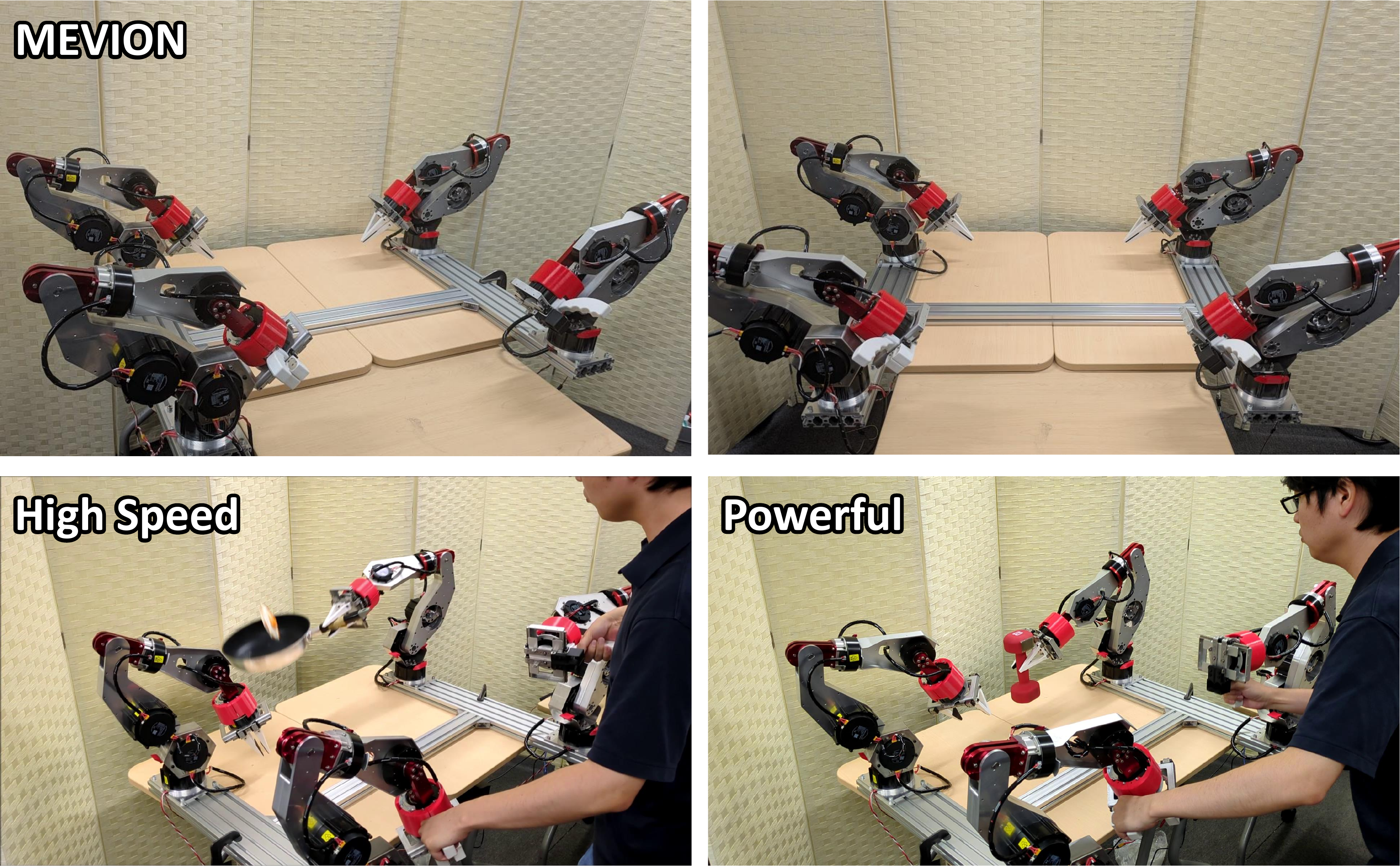}
  \vspace{-1.0ex}
  \caption{MEVION: Low-cost open-source data collection system for powerful and high-speed dual-arm manipulation, developed in this study.}
  \label{figure:mevion}
  \vspace{-3.0ex}
\end{figure}

\begin{figure*}[t]
  \centering
  \includegraphics[width=1.7\columnwidth]{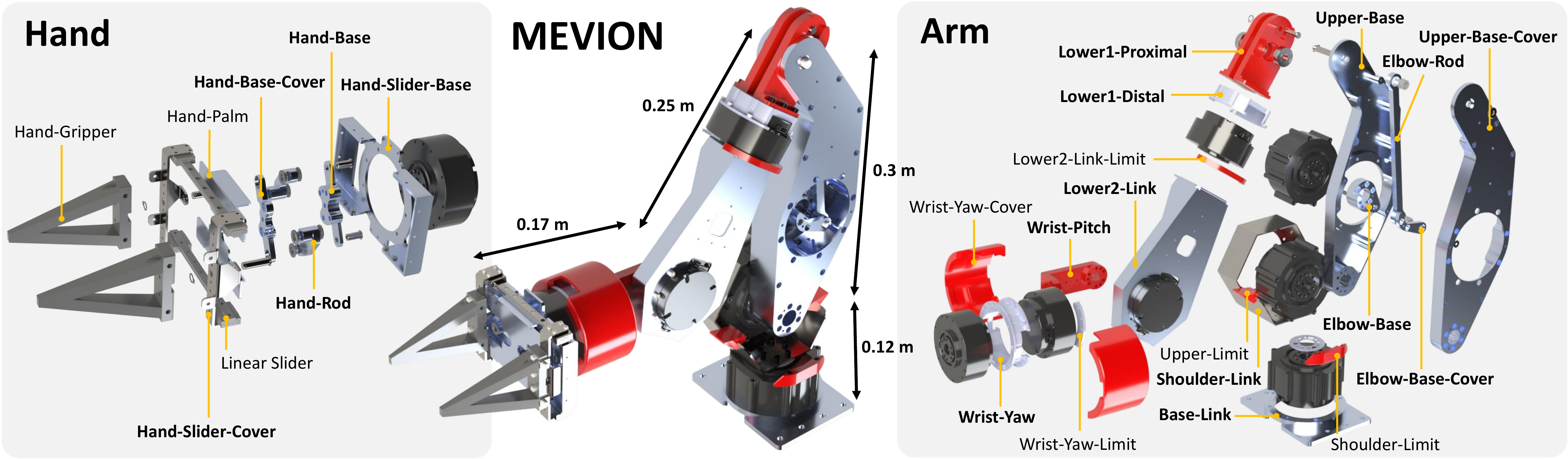}
  \vspace{-1.0ex}
  \caption{Design overview of MEVION: The design is divided into the arm and hand, comprising 17 essential metal components in total --- 11 in the arm and 6 in the hand, three of which are fabricated through sheet metal welding to form single integrated parts with complex and large geometries.
  }
  \label{figure:mevion-design}
  \vspace{-3.0ex}
\end{figure*}

\switchlanguage%
{%
  To address these limitations, we developed MEVION, a low-cost and open-source dual-arm robot data collection system capable of generating greater force and speed.
  All components of this robot can be sourced through e-commerce, and by leveraging sheet metal welding, a large body structure can be constructed with a small number of parts at low cost.
  MEVION is equipped with four 6-DoF arms with parallel grippers.
  Each arm weighs 7.0 kg and provides a maximum torque of 60 Nm, and the entire system can be built for approximately USD 14,000.
  The elbow joint adopts a closed-link mechanism similar to those used in quadruped robots, reducing the distal mass and enabling higher speed and force output at the end-effector.
  With MEVION, it becomes possible to manipulate heavy objects and perform high-speed operations, enabling the collection of more diverse manipulation data.
  Furthermore, we demonstrated a complete imitation learning pipeline with MEVION, confirming its capability for autonomous motion generation.
  All hardware and software developed in this study are included in the Supplementary Materials or \href{https://github.com/haraduka/mevion}{\textcolor{magenta}{github.com/haraduka/mevion}}.
}%
{%
  そこで本研究では, より大きな力と速度を出すことができる低コストかつオープンソースな双腕ロボットのデータ収集システムとして, MEVIONを開発した.
  本ロボットは全ての部品をE-Commerceにより調達可能であり, 板金溶接を駆使して大きな身体構造を少ない部品数で安価に構築している.
  MEVIONは6自由度+平行グリッパのアームを4つ備え, 各アームの重量は7.0kg, 最大トルクは60Nmであり, 全部で約14kドルで構築することができる.
  MEVIONの肘関節は四脚ロボットに見る閉リンク構造を採用しており, 手先の重量を軽減し, より大きな速度と力を出すことができる.
  このMEVIONにより, 重量物体の操作や高速な物体操作が可能となり, より多様なマニピュレーションデータの収集が期待できる.
  また, 摸倣学習の一連の流れを実際に行い, 自律的な動作生成能力について確認した.
  本研究の全てのハードウェア・ソフトウェアはSupplementary Materialsに含まれている (採択後, GitHubにも公開する予定である).
}%

\begin{figure}[t]
  \centering
  \includegraphics[width=0.7\columnwidth]{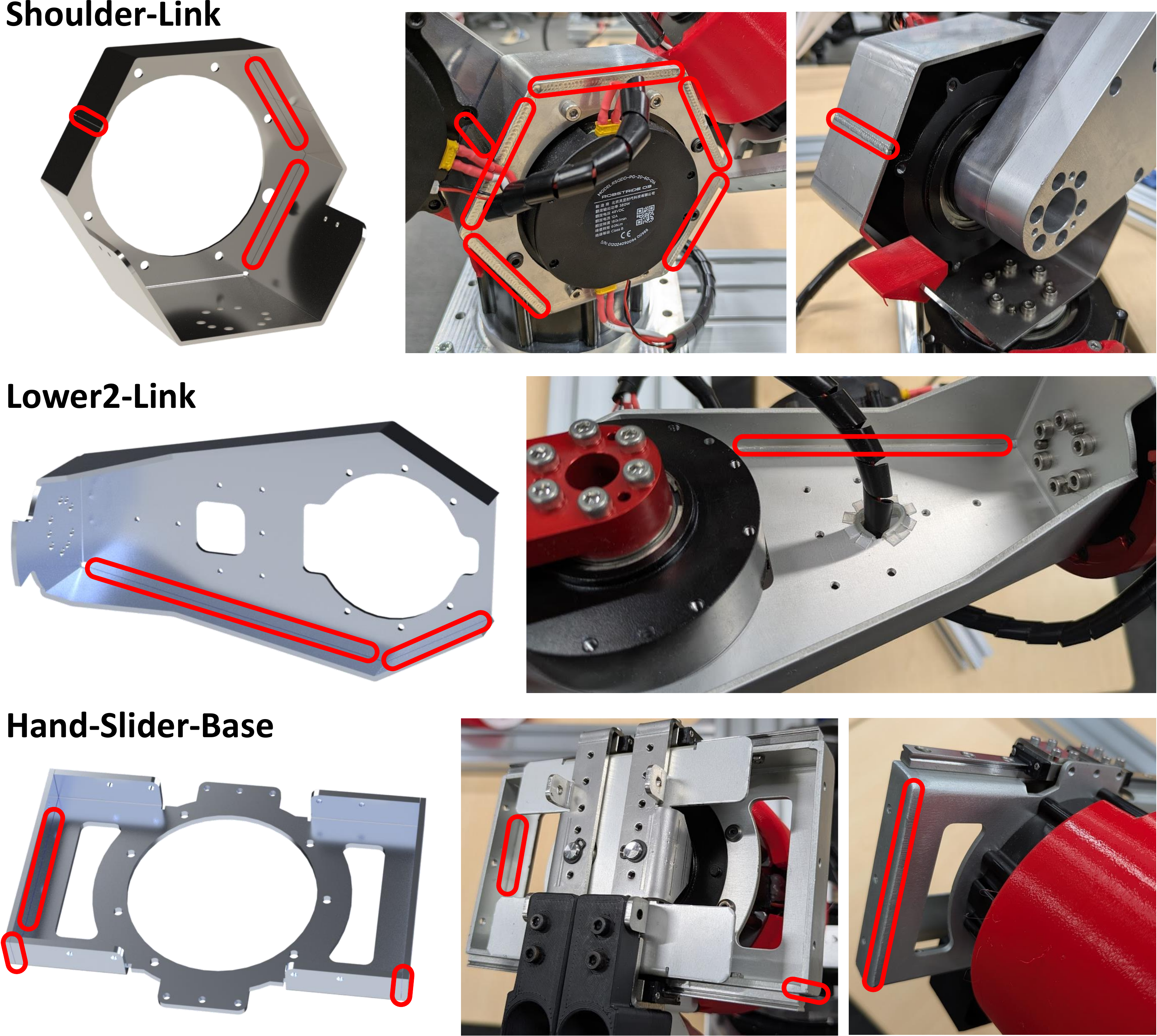}
  \vspace{-1.0ex}
  \caption{Details of sheet metal welding for the Shoulder-Link, Lower2-Link, and Hand-Slider-Base. The welded sections are highlighted in red.}
  \label{figure:mevion-welding}
  \vspace{-3.0ex}
\end{figure}

\section{Hardware and Software of MEVION} \label{sec:mevion}

\subsection{Design Overview of MEVION} \label{subsec:mevion-design}
\switchlanguage%
{%
  The design overview of MEVION is shown in \figref{figure:mevion-design}.
  The design is primarily divided into the arm and hand components, with the essential metallic structural parts comprising 11 components for the arm and 6 for the hand, totaling 17 components.
  When including joint angle limiters and cover parts, the total number of metallic components amounts to 21.
  For materials, all machined metal components are made of A7075 aluminum alloy.
  Among the sheet metal components, only the Shoulder-Link and Hand-Slider-Cover are made of SUS304 stainless steel, while the rest use A5052 aluminum alloy.
  In addition, some joint angle limiters and cable covers are fabricated in flexible TPU, while the fingertip structures of the leader and follower hands are made of PLA, both manufactured using 3D printing.

  The arm's joint configuration is identical to that of ALOHA, with six degrees of freedom consisting of Shoulder-Yaw, Shoulder-Pitch, Elbow-Pitch, Elbow-Yaw, Wrist-Pitch, and Wrist-Yaw.
  In general, each joint is driven by a single motor; however, for the elbow joint, a parallel-link mechanism, similar to those used in quadruped robots \cite{kawaharazuka2024mevius, kawaharazuka2026mevius2}, is employed to drive the joint from a proximal motor, thereby reducing the distal mass.
  The hand adopts a slider-crank parallel gripper mechanism, where a single motor simultaneously drives two fingers.
  The motors used are: RobStride03 for Shoulder-Yaw, Shoulder-Pitch, and Elbow-Pitch; RobStride02 for Elbow-Yaw, Wrist-Pitch, and Wrist-Yaw; and RobStride01 for the hand.
  Their respective maximum torques are 60 Nm, 17 Nm, and 17 Nm.

  A notable feature of MEVION is its small number of parts, achieved through extensive use of sheet metal welding, as shown in \figref{figure:mevion-welding}.
  The Shoulder-Link and Lower2-Link of the arm, as well as the Hand-Slider-Base of the hand, integrate large and complex shapes into single units via sheet metal welding, while the remaining parts are manufactured through machining.
  By leveraging sheet metal welding that provides sufficient structural strength even for large-scale shapes, a large body structure is realized with a small number of components, achieving both low cost and ease of assembly.
  All of these metallic components are designed to be fully compatible with meviy \cite{misumi2024meviy}, MISUMI's online machining and fabrication service, which automatically generates quotations and orders from 3D CAD files.
  As a result, the entire ordering process can be completed through e-commerce, enabling MEVION to be easily reproduced.
}%
{%
  MEVIONの設計概要を\figref{figure:mevion-design}に示す.
  MEVIONの設計は主にアーム部分とハンド部分に分かれており, 身体構造を成す必須の金属品については, アーム部分が11部品, ハンド部分が6部品の計17部品で構成されている.
  ここに関節角度リミット部品やカバー部品などを加えると, 金属部品は合計21部品となる.
  素材としては, 金属切削部品は全てA7075, 板金部品・板金溶接部品はShoulder-LinkとHand-Slider-CoverのみSUS304, それ以外はA5052を用いている.
  その他, 一部の関節角度リミット部品, ケーブルカバーなどは柔軟なTPUで, LeaderとFollowerのハンドの手先構造はPLAで, 3Dプリンタを用いて製作されている.

  アームの関節構造はALOHAと全く同じであり, 関節は順にShoulder-Yaw, Shoulder-Pitch, Elbow-Pitch, Elbow-Yaw, Wrist-Pitch, Wrist-Yawで6自由度となっている.
  基本的には単純に関節をモータ一つで駆動する方式を採用しているが, 肘の関節のみ, 四脚ロボットで見るような平行リンクを活用して近位のモータから関節を動かすことで, 手先の重さを軽減している.
  また, ハンドは一つのモータで2本の指を同時に駆動するスライダークランク機構の平行グリッパを採用している.
  用いているモータはShoulder-Yaw, Shoulder-Pitch, Elbow-PitchがRobStride03, Elbow-Yaw, Wrist-Pitch, Wrist-YawがRobStride02, ハンドがRobStride01であり, それぞれ最大トルクは60Nm, 17Nm, 17Nmである.
  RobStride01以外はエンコーダが2つついており, 絶対角を取得可能である.
  なお, 実際にはこれが4つ配置され, \figref{figure:mevion}のようなLeader-Followerの構成となっている.

  MEVIONの特筆すべき特徴は, その部品点数の少なさとそれを実現するための板金溶接の活用である.
  MEVIONに用いられている全ての金属部品を\figref{figure:mevion-parts}に示す.
  また, これに一部活用されている板金溶接の詳細を\figref{figure:mevion-welding}に示す.
  アーム部品のShoulder-LinkとLower2-Link, ハンド部品のHand-Slider-Baseは, 板金溶接により複雑かつ大きな形状を一体化しており, それ以外の部品は切削加工により製作されている.
  大きな形状でも強度が出せる板金溶接を活用することで, 少ない部品数で大きな身体構造を実現し, 安い価格とアセンブリの容易さを実現した.
  そして, これら全ての金属部品はMISUMIの切削・加工サービスであるmeviy \cite{misumi2024meviy}により, 3D形状ファイルから自動的に見積り・発注することが可能な設計となっている.
  ゆえに, 部品発注の全ての操作をE-Commerceで完結させることができ, 簡単にMEVIONを再現することができる.
}%

\begin{table*}[t]
  \centering
  \caption{Comparison between existing open-source leader-follower bimanual data collection systems and MEVION}
  \vspace{-3.0ex}
  \begin{tabular}{l|ccccccccc}
    \multirow{2}{*}{Name} & \multirow{2}{*}{Morphology} & \multirow{2}{*}{DOFs} & \multirow{2}{*}{Weight} & \multirow{2}{*}{Length\textsuperscript{*1}} & \multicolumn{2}{c}{Materials} & \multirow{2}{*}{Maximum Torque\textsuperscript{*2}} & \multirow{2}{*}{Maximum Speed\textsuperscript{*2}} & \multirow{2}{*}{Price\textsuperscript{*3}} \\
    & & & & & Arm & Hand & & \\ \hline
    SO-101 \cite{huggingface2025so101} & Tabletop & 4+1 & 0.7 kg & 0.45 m & Plastic        & Plastic        & 1.9 Nm  & 5.5 rad/s  & \$120\\
    ALOHA \cite{zhao2023aloha}         & Tabletop & 6+1 & 4.3 kg & 0.75 m & \textbf{Metal} & Plastic        & 21.2 Nm & 3.1 rad/s  & \$4,900\\
    OpenArm \cite{enactic2025openarm}  & Standing & 7+1 & 5.5 kg & 0.63 m & \textbf{Metal} & \textbf{Metal} & 40 Nm   & 16.8 rad/s & \$3,200\\
    MEVION (this study)                & Tabletop & 6+1 & 7.0 kg & 0.83 m & \textbf{Metal} & \textbf{Metal} & 60 Nm   & 20.4 rad/s & \$3,500\\
  \end{tabular}
  \label{table:comparison}
  \vspace{-1.0ex}
  \begin{flushleft}
  \footnotesize
    \textsuperscript{*1}The distance from the center of the first joint to the end-effector;
    \textsuperscript{*2}The peak torque and no-load joint speed of the highest-output motor;
    \textsuperscript{*3}The approximate cost of a single arm.
  \end{flushleft}
  \vspace{-6.0ex}
\end{table*}

\begin{figure}[t]
  \centering
  \includegraphics[width=0.8\columnwidth]{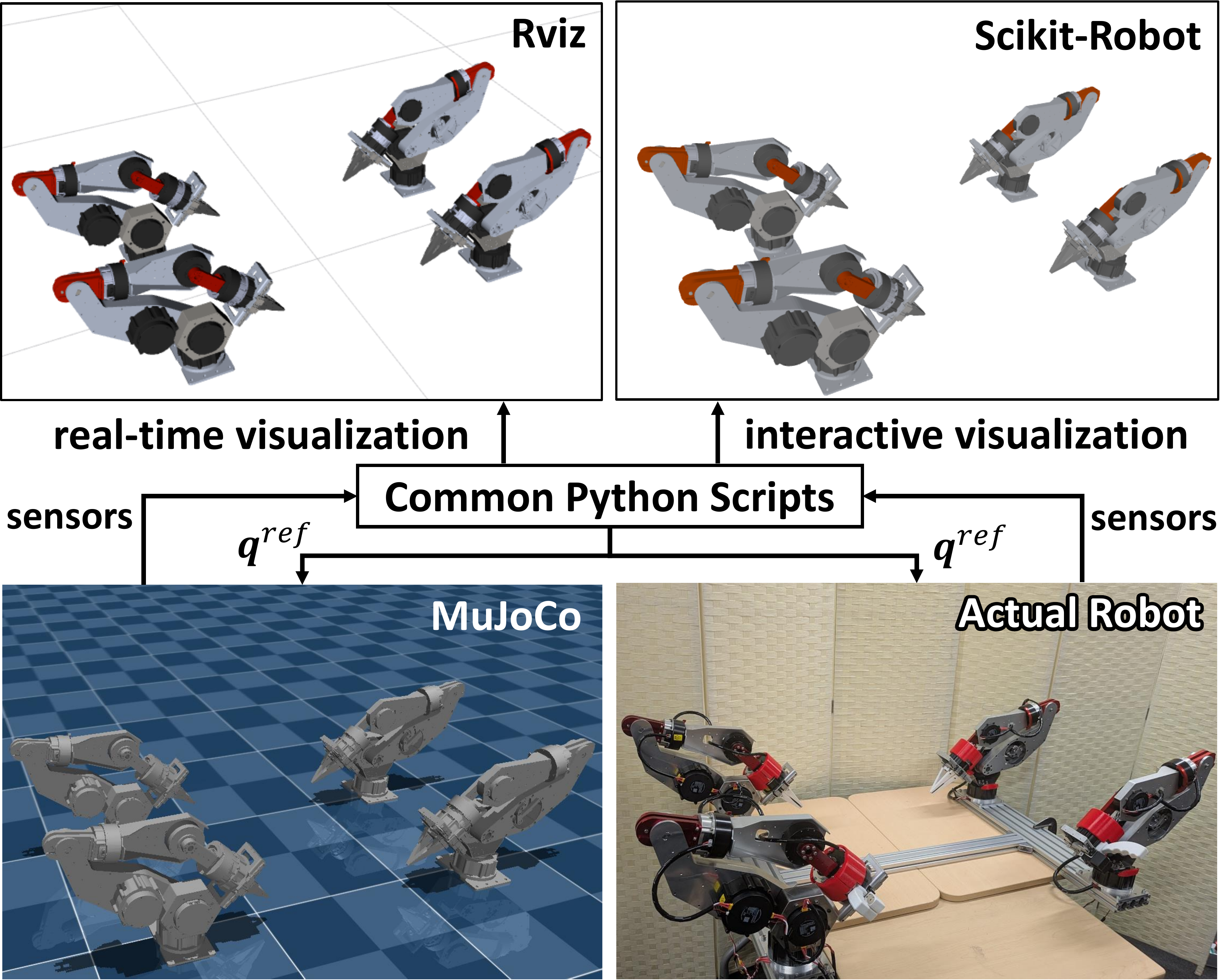}
  \vspace{-1.0ex}
  \caption{Software of MEVION: MEVION is controlled by a unified Python script, allowing both simulation on MuJoCo and real-world control to be executed with the same code. It supports real-time visualization of the robot state in RViz via ROS, as well as interactive visualization prior to sending control commands using Scikit-Robot \cite{yanokura2025scikitrobot}.}
  \label{figure:mevion-software}
  \vspace{-3.0ex}
\end{figure}

\subsection{Comparison with Existing Robot Systems} \label{subsec:comparison}
\switchlanguage%
{%
  The comparison between MEVION and existing open-source leader-follower dual-arm robot data collection systems is shown in \tabref{table:comparison}.
  Eight aspects were considered: morphology, number of DOFs, weight, length, materials, maximum torque, maximum speed, and price, comparing representative systems SO-101 \cite{huggingface2025so101}, ALOHA \cite{zhao2023aloha}, OpenArm \cite{enactic2025openarm}, and MEVION.
  Among them, SO-101, ALOHA, and MEVION are tabletop-type systems, whereas OpenArm is standing-type.

  Comparing the three tabletop-type systems, weight, length, and maximum torque increase stepwise in the order SO-101 $<$ ALOHA $<$ MEVION.
  In other words, these three robots clearly differ in terms of overall size and output capability.
  Regarding materials, SO-101's structural components are all plastic, ALOHA uses metal only in the arms, and MEVION uses all-metal components, reflecting a progressive increase in structural robustness.
  Notably, compared with ALOHA, MEVION is about 1.6 times heavier, yet achieves approximately three times the maximum torque and seven times the maximum speed, enabling significantly greater force and speed.
  Unlike SO-101 and ALOHA, MEVION places the elbow motor proximally and employs a parallel link mechanism, further raising the upper limits of torque and speed.
  Furthermore, MEVION remains low-cost, at around USD 3,500 per arm, while offering high performance.
  All of these systems are open-source and thus freely reconfigurable, and both SO-101 and ALOHA also have wheeled mobile-base variants \cite{fu2024mobile}.

  Next, we compare MEVION with OpenArm, which has relatively similar specifications.
  While MEVION is tabletop-type and OpenArm is standing-type, leading to substantial differences in joint sequence and count that preclude direct comparison, their weight, length, maximum torque, maximum speed, and price are all very close.
  The major differences lie in end-effector weight reduction and the number of parts.
  As described earlier, MEVION reduces end-effector weight and required gravity-compensation torque by arranging the elbow motor proximally.
  The peak payload at maximum extension is 7.6 kg for MEVION and 4.5 kg for OpenArm, indicating that MEVION can handle heavier loads.
  In addition, OpenArm consists of 31 basic metallic components, whereas MEVION uses only 17 (21 including joint limiters and covers), representing a substantial reduction in the number of parts.
  This is enabled by sheet metal welding, which allows integration of large and complex shapes while maintaining strength.
}%
{%
  既存のオープンソースのLeader-Follower方式の双腕ロボットデータ収集システムとMEVIONを比較した結果を\tabref{table:comparison}に示す.
  ここでは, morphology, the number of DOFs, weight, length, materials, maximum torque, maximum speed, and priceの8つの観点で, 代表的なSO-101 \cite{huggingface2025so101}, ALOHA \cite{zhao2023aloha}, OpenArm \cite{enactic2025openarm}とMEVIONの比較を行った.
  SO-101, ALOHA, MEVIONはテーブルトップ型であり, OpenArmは立位型である.

  まず, テーブルトップ型の3つのシステムを比べると, weight, length, maximum torqueはSO-101 $<$ ALOHA $<$ MEVIONの順で段階的に大きくなっている.
  つまり, この3つは異なるロボットとしての大きさと出力を持っていることがわかる.
  素材についても, SO-101は構造部品は全てプラスチック, ALOHAはアームのみ金属, MEVIONは全て金属と, 段階的に頑健な構造となっている.
  特に, MEVIONはALOHAよりも, 重さは約1.6倍ながら, 最大トルクは約3倍, 最大速度は約7倍と, より大きな力と速度を出すことができる.
  MEVIONはSO-101やALOHAと違い, 肘のモータを近位に配置し平行リンクを活用していることからも, トルクや速度の上限をさらに引き上げることができている.
  また, MEVIONは一本の腕で価格も約3,500ドルと, 低コストかつ高性能なシステムとなっている.
  なお, これらはどれもオープンソースであり自由に構造を変化させることができ, 特に特にSO-101とALOHAは車輪がついた台車型の構成も存在している\cite{fu2024mobile}.

  次に, このMEVIONに, 比較的近いスペックのOpenArmと比較する.
  MEVIONはテーブルトップ型なのに対して, OpenArmは立位型である.
  そのため, 関節の順番や数は大きく異なり単純な比較は難しいが, weight, length, maximum torque, maximum speed, priceは非常に近い値となっている.
  その一方で, 大きく異なるのは手先の軽量化と部品点数である.
  前述の通り, MEVIONは肘関節のモータを近位に配置することで, 手先を軽量化し, 必要な重力補償トルクを軽減することができる.
  そのため, MEVIONの最大伸展時のpeak payloadは7.6 kg, OpenArmは4.5 kgと, MEVIONの方がより大きな重量物を扱うことができる.
  また, OpenArmの部品点数が基本的な金属部品で31なのに対して, MEVIONは基本的な金属部品で17, 関節角度リミットやカバーを入れても21と, 非常に少ない部品数で構築されている.
  これは, 板金溶接の活用により, 強度を保った形での大きく複雑な形状の一体化が可能となっているためである.
  また, 他のロボットのハンドは一つのリニアスライダーのみを活用しているが, MEVIONのハンドは2本のスライダーを活用し, より頑健な作りとなっている.
}%

\begin{figure}[t]
  \centering
  \includegraphics[width=0.87\columnwidth]{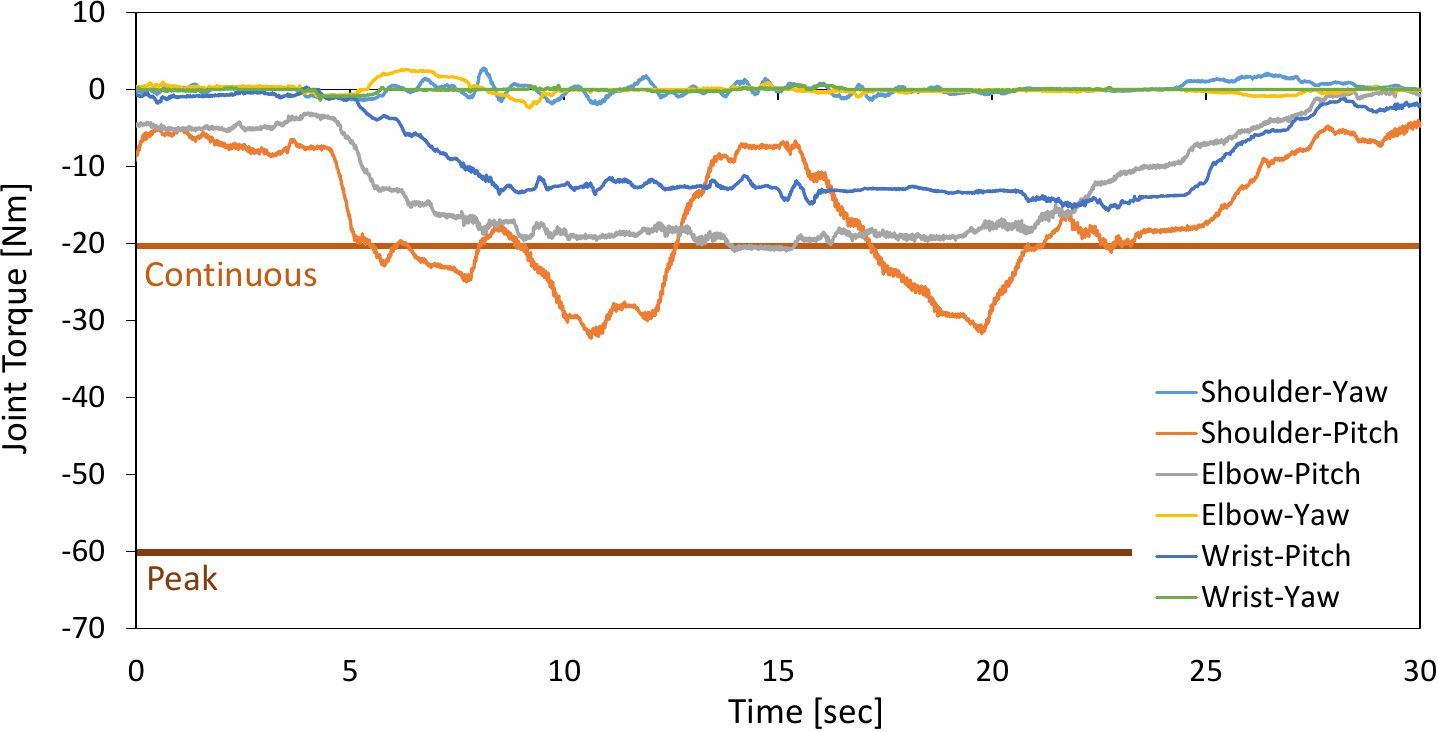}
  \vspace{-1.0ex}
  \caption{The transition of joint torques during the dumbbell manipulation.}
  \label{figure:dumbbell-torque}
  \vspace{-1.0ex}
\end{figure}

\begin{figure}[t]
  \centering
  \includegraphics[width=0.87\columnwidth]{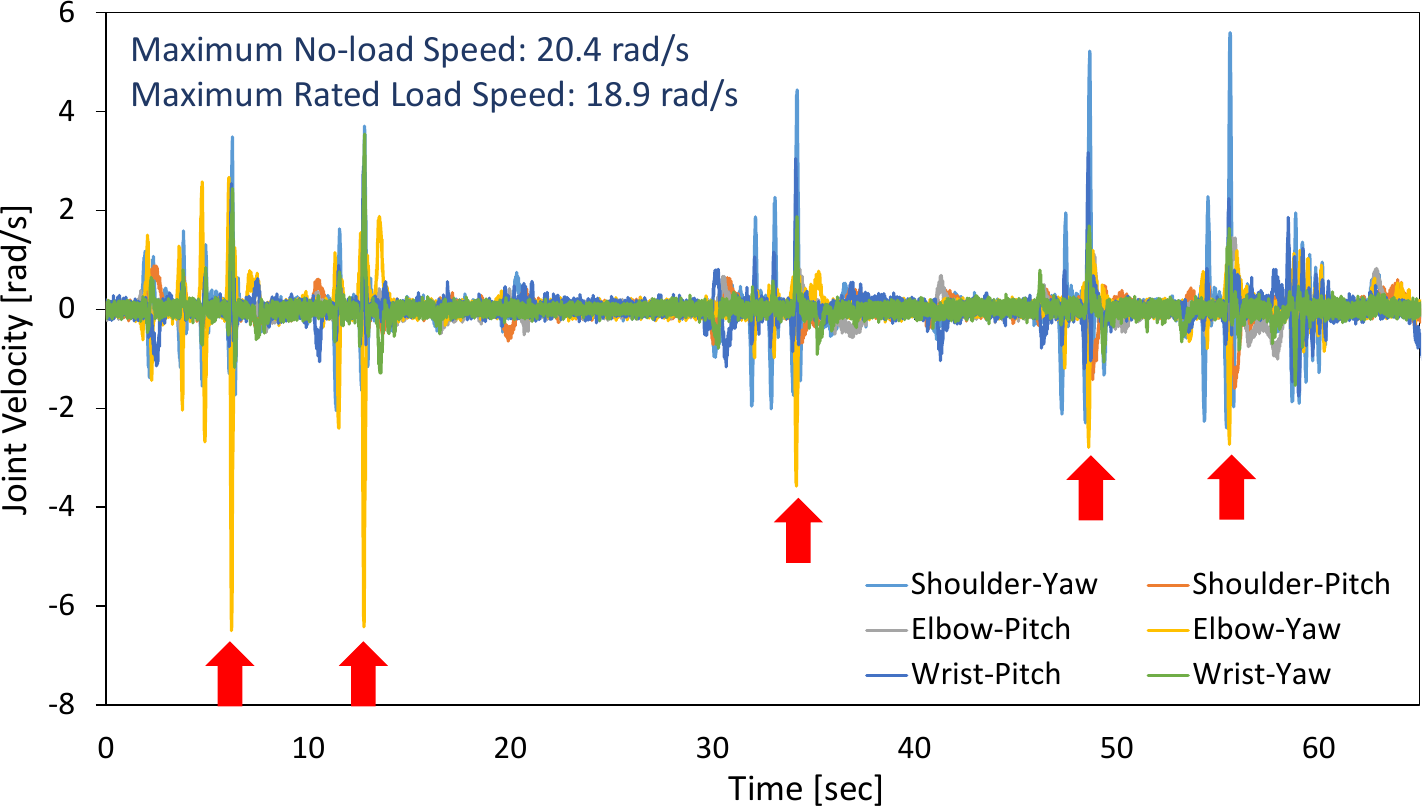}
  \vspace{-1.0ex}
  \caption{The transition of joint velocities during the Daruma Otoshi.}
  \label{figure:daruma-velocity}
  \vspace{-3.0ex}
\end{figure}

\begin{figure*}[t]
  \centering
  \includegraphics[width=1.65\columnwidth]{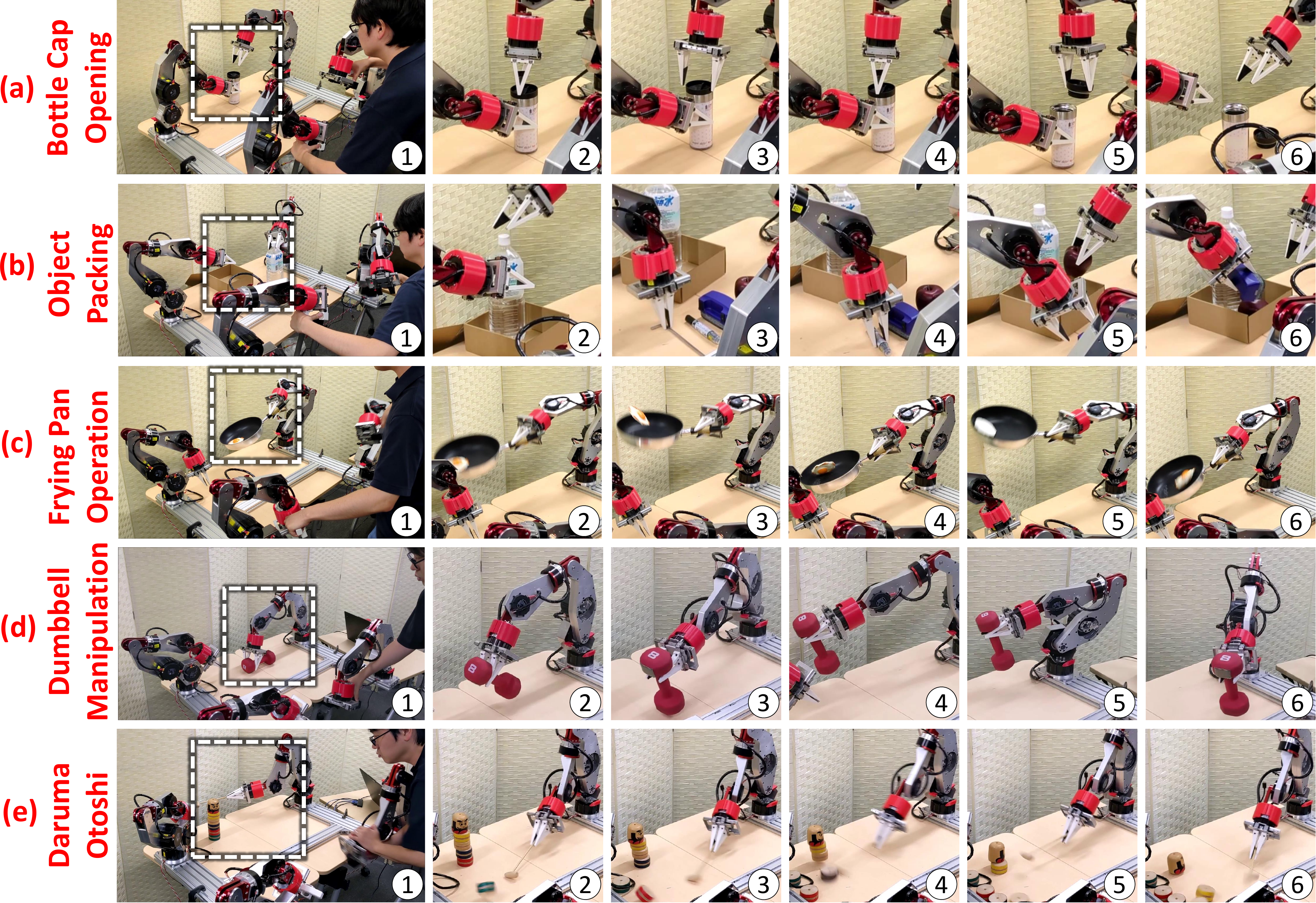}
  \vspace{-1.0ex}
  \caption{Teleoperation experiments using MEVION: (a) bottle cap opening, (b) object packing, (c) frying pan operation, (d) 3.6 kg dumbbell manipulation, and (e) Daruma Otoshi.}
  \label{figure:teleop-exp}
  \vspace{-1.0ex}
\end{figure*}

\begin{figure*}[t]
  \centering
  \includegraphics[width=1.65\columnwidth]{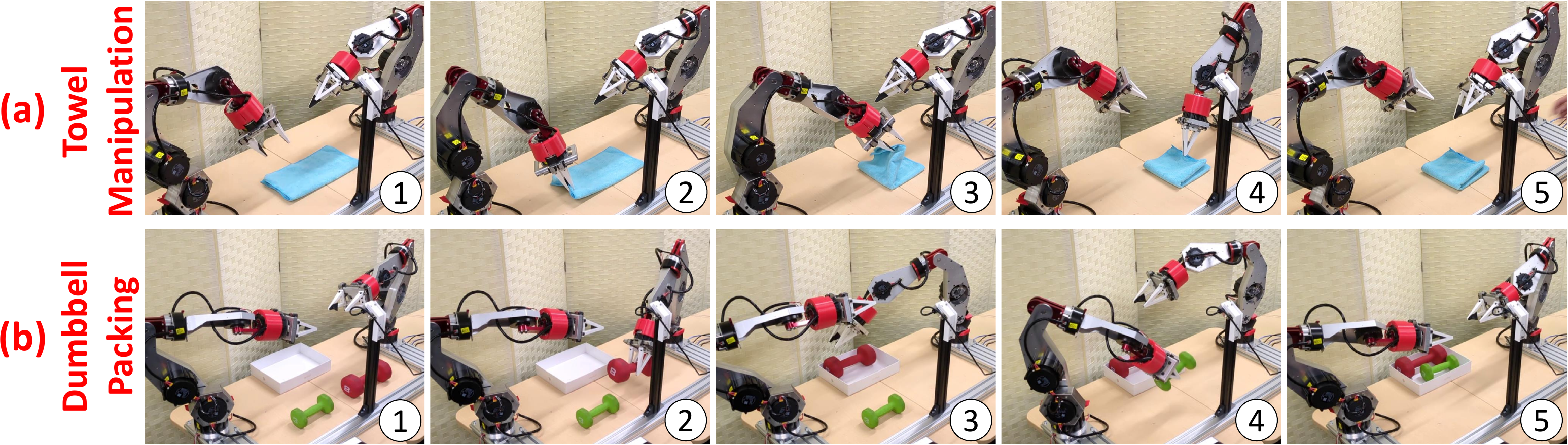}
  \vspace{-1.0ex}
  \caption{The results of imitation learning using MEVION: (a) towel manipulation and (b) dumbbell packing.}
  \label{figure:imitation-exp}
  \vspace{-3.0ex}
\end{figure*}

\subsection{Software of MEVION} \label{subsec:mevion-software}
\switchlanguage%
{%
  The software architecture of MEVION is shown in \figref{figure:mevion-software}.
  MEVION is primarily controlled through a unified Python script, enabling both MuJoCo-based simulation and real-world control to be executed with the same code.
  It supports real-time visualization of the robot's state in RViz via ROS, as well as interactive rendering before sending control commands using Scikit-Robot \cite{yanokura2025scikitrobot}.
  Unless otherwise specified, Intel RealSense D405 cameras are employed as both fixed and wrist-mounted cameras.

  MEVION operates on either 24V or 48V power and adopts CAN communication. Four CAN-USB interfaces are connected to the individual arms, which are then linked to a PC for control.
  The basic control scheme is a PD controller that computes current commands based on joint angle commands, with gravity compensation torque added.
}%
{%
  MEVIONのソフトウェア構成を\figref{figure:mevion-software}に示す.
  MEVIONは基本的に共通のPythonスクリプトで制御されており, MuJoCo上でのシミュレーションと実機の制御を同じコードで行うことができる.
  ROSを用いたRVizにおける実機状況のリアルタイム描画と, Scikit-Robot \cite{yanokura2025scikitrobot}を用いた制御指令を送る前のインタラクティブな描画の両方をサポートしている.
  カメラは特に指定がなければIntel RealSense D405を固定カメラと手先カメラとして用いている.

  MEVIONは24Vまたは48Vで駆動され, 通信方式はCAN通信を採用している.
  4つのCAN-USBインターフェイスが各アームに接続されており, これらをPCに接続して制御を行う.
  基本的な制御方法は, モータの関節角度指令に対する電流指令のPD制御であり, 重力補償トルクが加算されている.
}%

\section{Experiments} \label{sec:experiment}


\subsection{Teleoperation Experiments}

\switchlanguage%
{%
  The teleoperation experiment results are shown in \figref{figure:teleop-exp}.
  Five tasks were performed: (a) bottle cap opening, (b) object packing, (c) frying pan operation, (d) 3.6 kg dumbbell manipulation, and (e) the traditional Japanese game Daruma Otoshi.
  These tasks range from (a) bimanual cooperative actions, to (b) the transportation of slender and heavy objects including wrenches and 1.5 kg plastic bottles, and to (c) high-speed manipulations using a frying pan, confirming that MEVION can execute a wide variety of tasks.
  The joint torques during (d) dumbbell manipulation are presented in \figref{figure:dumbbell-torque}, and the joint velocities during (e) Daruma Otoshi are presented in \figref{figure:daruma-velocity}.
  In the dumbbell task, all joints produced torque values with sufficient margin from their maximum limits.
  In the Daruma Otoshi task, the maximum joint velocity reached 6.5 rad/s, demonstrating that MEVION enables high-speed manipulation that would be difficult with systems such as ALOHA or SO-101.
}%
{%
  テレオペレーションの実験結果を\figref{figure:teleop-exp}に示す.
  ここでは, ボトルキャップ開け, 様々な物体のパッキング, フライパンによる調理, 3.6kgのダンベル操作, だるま落としの5つのタスクを行った.
  ボトルキャップ開けのような双腕協調動作から, レンチやペットボトルを含む細い物体や重量物体の移動操作, フライパンを用いた高速な物体操作まで, 多様なタスクを実行できることを確認した.
  また, ダンベル操作時の各関節のトルクを\figref{figure:dumbbell-torque}に, だるま落とし時の各関節の速度を\figref{figure:daruma-velocity}に示す.
  ダンベル操作について, すべての関節において, 最大値から余裕を持ってトルクが出せていることがわかる.
  まただるま落としについて, 最大の関節速度は6.5rad/sに達しており, ALOHAやSO-101では難しい高速な物体操作が可能であることがわかる.
}%

\subsection{Imitation Learning Experiments}

\switchlanguage%
{%
  The inference results of imitation learning are shown in \figref{figure:imitation-exp}.
  Two tasks were performed: (a) folding a towel and placing it in the center, and (b) packing dumbbells weighing 3.6 kg and 1.4 kg into a box.
  First, a human operator collected 15 demonstrations for each task via teleoperation.
  Using the collected data, imitation learning was conducted with the Action Chunking Transformer \cite{zhao2023aloha}.
  For both tasks, we confirmed that the robot successfully executed autonomous actions more than five times consecutively, even when the positions of the towel or dumbbells were slightly shifted.
}%
{%
  実際にデータ収集から摸倣学習, 推論までの一連の流れを行った際の結果を\figref{figure:imitation-exp}に示す.
  タオルを畳んで中央に配置するタスクと, 3.6 kgと1.4 kgのダンベルを箱に詰めるタスクを行った.
  まずは人間がテレオペレーションにより15回ずつデータ収集を行い, 収集したデータを用いてAction Chunking Transformer \cite{zhao2023aloha}により摸倣学習を行った.
  どちらのタスクも, タオルやダンベルの位置をずらしても, 連続して5回以上の自律動作に成功することを確認した.
}%

\section{CONCLUSION} \label{sec:conclusion}
\switchlanguage%
{%
  In this work, we developed MEVION, a low-cost and open-source tabletop dual-arm data collection platform that achieves greater force and speed than existing systems.
  All components of MEVION can be sourced via e-commerce, particularly through MISUMI and meviy, and its large body structure is realized with a small number of parts at low cost by leveraging sheet metal welding.
  With MEVION, it becomes possible to manipulate heavy objects and perform high-speed operations, enabling the collection of more diverse manipulation data.
  We expect that this open-source hardware and software will be widely adopted by researchers and contribute to advancing dual-arm manipulation as well as education in the field.
}%
{%
  本研究では, テーブルトップの双腕動作データ収集のプラットフォームとして, 既存のシステムよりも大きな力と速度を出すことができる, 低コストかつオープンソースなMEVIONを開発した.
  MEVIONは全ての部品をE-Commerce, 特にMISUMIとmeviyにより調達可能であり, 板金溶接を駆使して大きな身体構造を少ない部品数かつ安価に構築することができている.
  実際に, MEVIONは6自由度+平行グリッパのアームを4つ備え, 各アームの重量は7.0kg, 最大トルクは60Nmであり, 全部で約14kドルで構築することができる.
  MEVIONを用いていることで, 重量物体の操作や高速な物体操作が可能となり, より多様なマニピュレーションデータの収集が可能であること, 摸倣学習の一連の流れを実際に行い, 自律的な動作生成が可能であることを確認した.
  今後, このオープンソースハードウェアが多くの研究者に活用され, より高度な双腕マニピュレーションや教育の発展に寄与することを期待している.
}%

{
  \bibliographystyle{IEEEtran}
  \bibliography{main}
}

\clearpage
\section*{Description of Supplementary Materials} \label{sec:appendix}

\subsection{Overview}
\switchlanguage%
{%
  The Supplementary Materials for this study include the following files:
  \begin{itemize}
    \item \textbf{PartsList.xlsx}: A list of all components used in MEVION.
    \item \textbf{MEVION-STEP.tar.gz}: STEP files for each individual MEVION component.
    \item \textbf{MEVION-Assembly.STEP}: The full assembly file of MEVION.
    \item \textbf{Software.tar.gz}: The software package for MEVION.
  \end{itemize}
}%
{%
  本研究のSupplementary Materialsに含まれるファイルは以下の通りである.
  \begin{itemize}
    \item PartsList.xlsx: MEVIONの部品リスト.
    \item MEVION-STEP.tar.gz: MEVIONの各部品のSTEPファイル.
    \item MEVION-Assembly.STEP: MEVIONのアセンブリファイル
    \item Software.tar.gz: MEVIONのソフトウェア.
  \end{itemize}
}%

\subsection{Hardware}
\begin{figure}[htb]
  \centering
  \includegraphics[width=0.8\columnwidth]{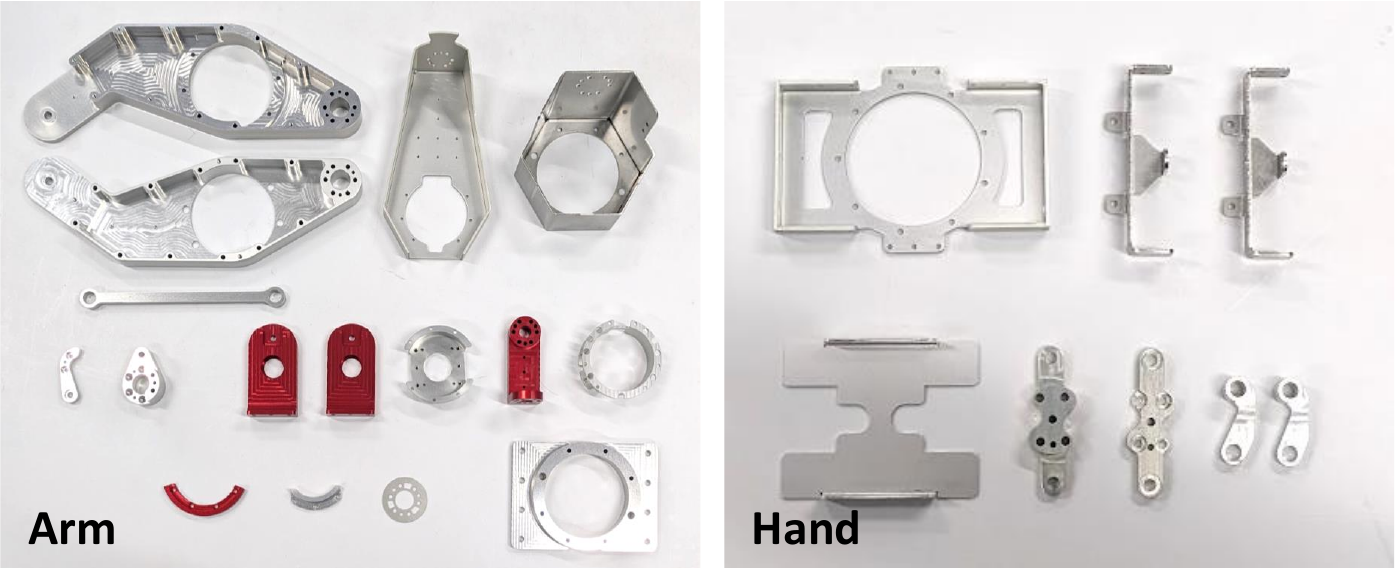}
  \vspace{-1.0ex}
  \caption{All metal components used in MEVION.}
  \label{figure:mevion-parts}
  \vspace{-1.0ex}
\end{figure}

\switchlanguage%
{%
  This section provides information related to the hardware, with a focus on key points regarding the parts list.
  Currently, the metallic parts of MEVION cost USD 1,465 for the arm and USD 421 for the hand, and including all motors and sensors, the total comes to USD 3,532.
  Since four sets are used, the entire system can be built for approximately USD 14,000.
  The parts list includes details such as part names, materials, quantities, and prices, but the most critical information is the part numbers used for ordering via meviy.
  All metal components used in this study can be ordered simply by specifying their part numbers.
  Of course, it is also possible to place orders by uploading the corresponding STEP files.
  All mechanical parts can be ordered through MISUMI and meviy, and for electronic components, individual URLs are provided in the list.
  Note that the parts list contains information for two types of hands: \textbf{hand} refers to a gripper driven by CyberGear, while \textbf{hand2} refers to a gripper driven by RobStride01.
  In addition, all components laid out are shown in \figref{figure:mevion-parts}.
  Assembly files for these components are provided, allowing users to assemble the system by referring to them.
  Owing to the small number of parts, the assembly process is straightforward, and once familiar with the procedure, it can be completed in one hour.
}%
{%
  ここではハードウェアに関する情報を提供する.
  特に, 部品リストについて主要なポイントを解説する.
  まず, 現状MEVIONの金属部品はアーム部分が1,465 USD, ハンド部分が421 USDであり, モータやセンサを全て入れると3,532 USDとなっている.
  これを4つ分用いるため, 合計して約14,000 USDで構築可能である.
  部品リストには名前, 素材, 個数, 価格などが書かれているが, 特に重要なのはmeviyで発注する部品の型番である.
  本研究の金属部品は全て, 型番を指定するのみで発注することができる.
  もちろん, STEPファイルをアップロードして発注しても良い.
  全ての機械部品がMISUMIとmeviyによって発注可能であり, 回路部品については個別にURLが記載されている.
  なお, 部品表には2種類のhandに関する部品情報が記載されているが, handはCyberGearを用いたグリッパ, hand2はRobStride01を用いたグリッパである.
  また, 全ての部品を並べたものを\figref{figure:mevion-parts}に示す.
  これら部品のアセンブリファイルを提供しているため, これを参照しながら組み立てることができる.
  部品点数が少ないこともあり, 組み立ては非常に簡単であり, 慣れれば1時間程度で完成させることができる.
}%

\subsection{Software}
\switchlanguage%
{%
  Here, we provide information on the software.
  All programs are written in Python and support both MuJoCo-based simulation and real-world control.
  In addition, visualization in RViz via ROS and interactive visualization using Scikit-Robot are also supported.
  Accordingly, the package includes XML model files for MuJoCo, URDF model files for RViz, and related resources.

  MEVION, unlike ALOHA, does not employ hardware-based gravity compensation; instead, gravity compensation torques are computed and applied on the software side.
  The control input for each joint $i$ is given by
  \begin{align}
    \tau^{cmd}_i
    = K_{p,i}(q_i^{ref}-q_i)
    + K_{d,i}(\dot q_i^{ref}-\dot q_i)
    + \tau^{ff}_i
  \end{align}
  where the feedforward term $\tau^{ff}$ is computed using Pinocchio \cite{carpentier2019pinocchio}.
  This enables smooth operation without dedicated gravity compensation mechanisms, as used in ALOHA.
  To ensure safe operation, limits are imposed on all of $q_i^{ref}$, $\dot q_i^{ref}$, and $\tau^{ff}_i$, corresponding to joint angle, joint velocity, and joint torque, respectively.
  The control loop runs at 200 Hz, with basic unilateral control implemented.

  While this approach eliminates the need for additional frames or mechanisms as used in ALOHA, it raises greater safety concerns.
  Therefore, safety mechanisms are incorporated both in hardware and software: a wireless emergency stop switch is implemented on the hardware side, while constraints such as maximum torque and velocity limits are enforced in software.
  In addition, sample programs for collision avoidance are provided.
}%
{%
  ここではソフトウェアに関する情報を提供する.
  全てのプログラムはPythonにより記述されており, MuJoCoによるシミュレーションと実機制御の両方をサポートしている.
  また, ROSによるRvizでの可視化, Scikit-Robotによるインタラクティブな可視化もサポートしている.
  そのため, MuJoCo用のXMLモデルファイル, Rviz用のURDFモデルファイルなども含まれている.

  MEVIONはALOHAとは異なり, ハードウェアで重力補償しておらず, ソフトウェア側で重力補償トルクを計算し, 適用している.
  各関節$i$の基本制御入力は
  \begin{align}
    \tau^{cmd}_i
    = K_{p,i}(q_i^{ref}-q_i)
    + K_{d,i}(\dot q_i^{ref}-\dot q_i)
    + \tau^{ff}_i
  \end{align}
  である．
  Pinocchio \cite{carpentier2019pinocchio}を用いて重力補償トルク$\tau^{ff}$が計算されており, ALOHAのように重力補償用の機構が無くとも滑らかに操作することができる.
  この際, $q_i^{ref}$, $\dot q_i^{ref}$, $\tau^{ff}_i$の全てについて, 関節角度・関節速度・関節トルクのリミットを設けることで, 安全に操作できるようにしている.
  制御周期は200Hzであり, 基本的なユニラテラル制御が実装されている.

  MEVIONはALOHAのように余計なフレームなどが必要ない反面, 安全性の面で懸念が多い.
  そのため, 無線緊急停止スイッチなどの安全機構をハードウェア実装, 最大トルク制約や最大速度制約などの安全機構をソフトウェア実装している.
  また, collision avoidanceに関するサンプルプログラムなども含まれている.
}%

\subsection{Limitations and Future Challenges}
\begin{figure}[htb]
  \centering
  \includegraphics[width=0.8\columnwidth]{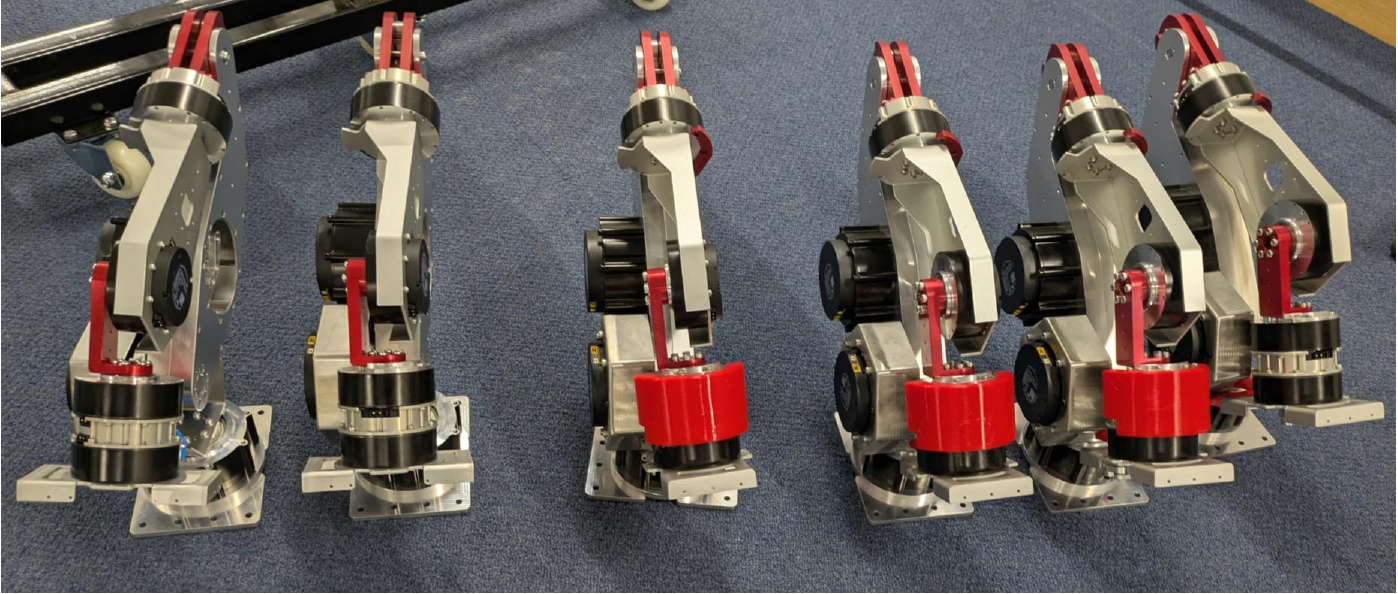}
  \vspace{-1.0ex}
  \caption{Researchers are currently building multiple MEVIONs for large-scale data collection.}
  \label{figure:massive-collection}
  \vspace{-1.0ex}
\end{figure}

\switchlanguage%
{%
  \textbf{Sheet metal welding}.
  While sheet metal welding is highly effective for reducing the number of components and overall cost, it also presents several challenges.
  First, unlike metal machining, it is difficult to perform modifications or repairs independently.
  Although re-ordering parts is often the simplest solution due to the relatively low cost, it may be necessary to consider the trade-offs with metal machining in future work.
  Second, compared to metal machining, the quality of welding can vary depending on the fabrication vendor.
  In leader-follower systems, such variability often does not pose a significant issue, as the robot is directly operated by a human.
  However, when scaling to large-scale data collection, it may be necessary to account for the impact of such variability.

  \textbf{Large-scale data collection}.
  This work ultimately aims to enable large-scale data collection.
  By performing gravity compensation in software, the physical footprint of the robot can be reduced, allowing a larger number of robots to be operated simultaneously.
  In practice, we are planning to construct multiple MEVION systems in collaboration with several researchers and utilize them for data collection (\figref{figure:massive-collection}).
  On the other hand, scalability introduces challenges, particularly in terms of mass production.
  It may be necessary to revisit the design with a focus on simplifying the structure and improving assembly efficiency to better support large-scale deployment.
}%
{%
  \textbf{板金溶接}.
  板金溶接は部品点数の削減やコスト削減に非常に有効な一方, いくつかの課題も存在する.
  まず, 金属切削と違い, 自分で加工したり修理したりすることが難しい点が挙げられる.
  価格が安いため新たに発注し直すことが最も簡単な解決策であるが, 今後金属切削とのトレードオフを考えていく必要があるかもしれない.
  次に, 板金溶接は金属切削と比べると, 業者によっては溶接の質にばらつきが出ることがある点が挙げられる.
  leader-follower systemの場合は人間がロボットを操作してしまうため, そのような質のばらつきが大きく問題にならないことが多いが, 今後大規模なデータ収集を行う上では, ばらつきの影響を考慮する必要があるかもしれない.

  \textbf{大規模データ収集}.
  本研究は, 最終的には大規模なデータ収集へ活用することを目的としている.
  重力補償をソフトウェア側で行うことで, ロボットの占有空間は小さくなるため, より多くのロボットを同時に動かすことが出来る点では利点がある.
  実際に, 一部研究者らと共同で, 複数のMEVIONを構築しデータ収集に活用する予定である(\figref{figure:massive-collection}).
  一方で, 大規模なデータ収集を行う上では, 特に量産性の問題が存在する.
  より簡易な構造設計や組み立てに関する工夫など, 大規模化に合わせた設計の見直しが必要になるかもしれない.
}%

\end{document}